\documentclass{article}

    \usepackage[preprint]{neurips_2024}

\usepackage[utf8]{inputenc} 
\usepackage[T1]{fontenc}    
\usepackage{hyperref}       
\usepackage{url}            
\usepackage{booktabs}       
\usepackage{amsfonts}       
\usepackage{nicefrac}       
\usepackage{microtype}      
\usepackage{xcolor}         
\usepackage{textcomp}
\usepackage{amsmath}
\usepackage{graphicx}
\usepackage{subcaption}
\usepackage{float}
\usepackage{titlesec}
\usepackage[english]{babel}
\usepackage{url}

\titlespacing*{\paragraph}{0pt}{0.4\baselineskip}{0.15\baselineskip}
\titlespacing*{\section}   {0pt}{0.8\baselineskip}{0.4\baselineskip}

\title{CLIP Embeddings for AI‑Generated Image Detection: A Few‑Shot Study with Lightweight Classfier}

\author{
  Ziyang Ou\\
  Department of Electrical and Computer Engineering\\
  University of Rochester\\
  Rochester, NY 14627, USA\\
  \texttt{zou9@ur.rochester.edu}\\
  }

\begin{document}

\maketitle

\begin{abstract}
Verifying the authenticity of AI-generated images presents a growing challenge on social media platforms these days. While vision-language models (VLMs) like CLIP outdo in multimodal representation, their capacity for AI-generated image classification is underexplored due to the absence of such labels during the pre-training process. This work investigates whether CLIP embeddings inherently contain information indicative of AI generation. 
A proposed pipeline extracts visual embeddings using a frozen CLIP model, feeds its embeddings to lightweight networks, and fine‑tunes only the final classifier. Experiments on the public CIFAKE benchmark show the performance reaches $\approx95\%$ accuracy without language reasoning. Few-shot adaptation to curated custom with $20\%$ of the data results in performance to $\approx85\%$. A closed‑source baseline (Gemini‑2.0) has the best zero‑shot accuracy yet fails on specific styles.
Notably, some specific image types, such as wide-angle photographs and oil paintings, pose significant challenges to classification. These results indicate previously unexplored difficulties in classifying certain types of AI-generated images, revealing new and more specific questions in this domain that are worth further investigation.
\end{abstract}

\section{Introduction}
The rapid growth of AI-generated image technologies, such as Generative Adversarial Networks (GANs) and Stable Diffusion, poses significant challenges to image authenticity. The widespread propagation of these fake images threatens social media and public trust. However, existing AI-generated content detection methods, including Convolutional Neural Networks (CNNs), often lack text comprehension capabilities.

This study is focused on two key questions: First, are the visual embeddings extracted by the CLIP model alone enough to separate real and AI-generated images? Second, is few-shot adaptation possible on a novel dataset? To address the questions above, a set of experiments was designed and experimented with. Initially, visual embeddings of real and AI-generated images were extracted using a pre-trained CLIP model. Embeddings were subsequently fed into a classification network, the logits output of which were used for binary classification. The purpose of these experiments is to assess the quality and discriminative power of visual embeddings and evaluate the potential and limitations of the embeddings.

Experiments were conducted on a publicly available dataset and a custom dataset. Notably, some categories of images, such as wide-angle and oil-painting-style images, posed a greater challenge to the proposed pipeline. Additionally, the same custom dataset was experimented with using a closed-source, prompt-based LLM (Google Gemini) to benchmark the capabilities of newer, gigantic-scale parameter models.

The remaining paper reviews related work from recent years in Section 2, describes specific experimental methods in Section 3, shows detailed results and findings in Section 4, and discusses and looks forward to future work in Section 5.

\section{Related work}

One of GAN fake image detection techniques suggested by \cite{1907.06515}, was frequency-domain-based for images. Their research showed that GAN models introduce artifacts into the frequency domain. These artificial patterns can serve as "fingerprints" for detection. Although visually realistic, GAN-generated images contain detectable frequency-domain patterns. GAN-synthesized images can be identified through spectrogram analysis and frequency domain feature extraction.

Recent research has demonstrated that Vision Transformers (ViTs) show promising performance on image semantic comprehension. In the work "Emerging Properties in Self-Supervised Vision Transformers", \cite{2104.14294} mention that self-supervised ViT features contain explicit information about the semantic segmentation of an image. It is a property not normally found in conventional ConvNetworks, suggesting the potential applicability of transformer-based architectures in different classification tasks.

A recent work by \cite{2402.12927} explores the effectiveness of pre-trained VLMs when paired with recent adaptation methods for universal deepfake detection. They keep both the visual and language encoders of CLIP active and apply fine‑tuning for transfer learning. By using just one‑third of the training images for training, their pre-tuned model improves on the SOTA approach by +5.01 mAP and +6.61 accuracy. Their research provides graphical proof of ViT linear separability boundaries and points out that CLIP embedding preservation can significantly enhance fake detection robustness.

CLIP embeddings already encode a "real‑versus‑fake" boundary. The work of \cite{2312.00195} shows that a simple CLIP-based detector achieves top performance on a huge variety of generators and in the most challenging settings. All the previous evidence has explained that CLIP already comes with a built-in true‑false separation boundary in its latent space.

\section{Method}

\subsection{CLIP Embedding + MLP / CNN on CIFAKE dataset}
\label{sec:clip-mlp}
Denote an RGB input image by $x\in\mathbb{R}^{H\times W\times3}$.
A frozen \textsc{CLIP} encoder $f_{\text{CLIP}}$ maps the image into a single 512-dimensional embedding,
\begin{equation}
    \mathbf z \;=\; f_{\text{CLIP}}(x)\in\mathbb{R}^{512}.
\end{equation}

\begin{figure}[H]
    \centering
    \begin{subfigure}[b]{0.43\linewidth}
        \centering
        \includegraphics[width=\linewidth]{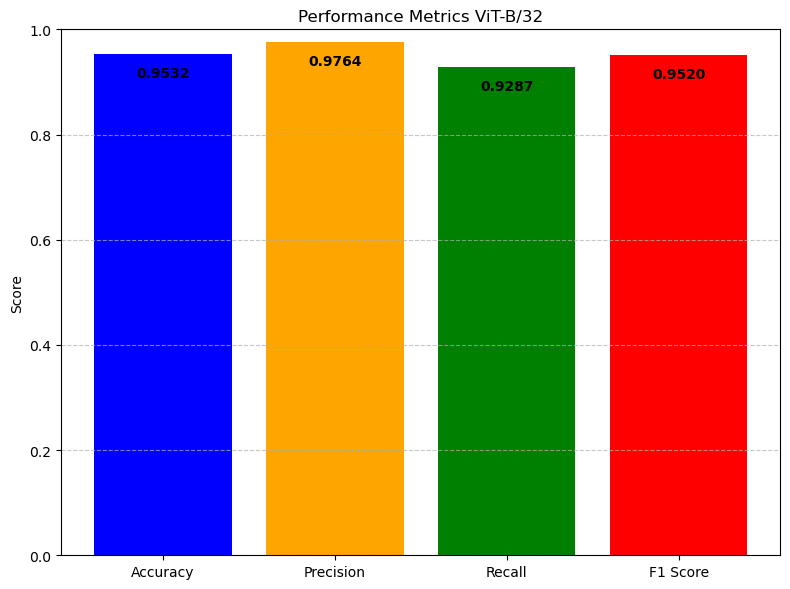}
        \caption{CLIP embeddings $\rightarrow$ MLP}\label{fig:cifake-mlp}
    \end{subfigure}
    \begin{subfigure}[b]{0.43\linewidth}
        \centering
        \includegraphics[width=\linewidth]{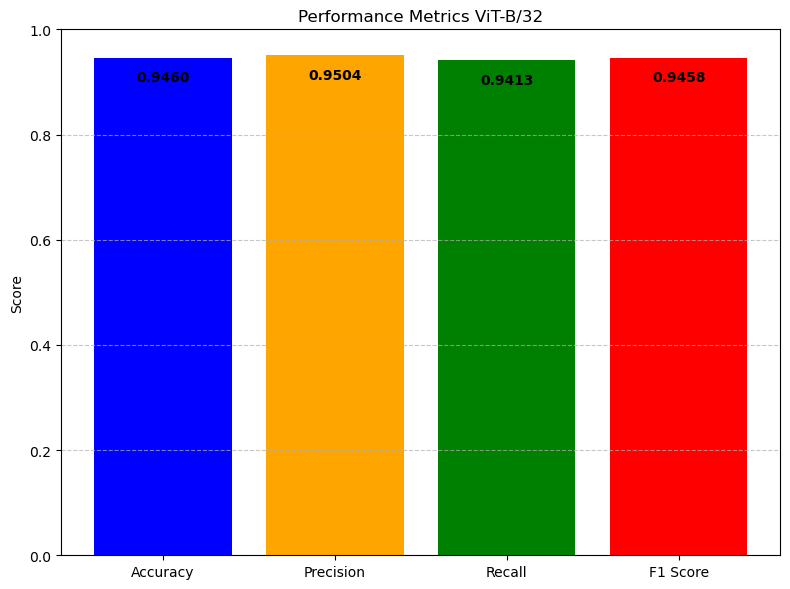}
        \caption{CLIP embeddings $\rightarrow$ ConvNet}\label{fig:custom-cnn}
    \end{subfigure}
    \caption{Performance metrics based on CIFAKE dataset. Achieves $\approx95\%$ Accuracy, Precision, and Recall on a large test set (12\,000 samples), far exceeding random guessing.}
    \label{fig:cifake_performance}
\end{figure}

To quantify how much information the frozen CLIP embedding $\mathbf z$ carries, two lightweight discriminators were trained and tested. Quantitative results are reported in Fig.~\ref{fig:cifake_performance}.

\emph{MLP.}
$\mathbf z \!\xrightarrow[]{512\to 256}\! \text{ReLU}\!\xrightarrow[]{} \text{Dropout}(0.2)\! \xrightarrow[]{128\to 1}\! \text{logit}$.

\emph{1‑D CNN head.}
Treating $\mathbf z$ as a one‐dimensional vector of length 512, a   
$\text{Conv1d}(1\!\to\!32,\text{k}=3)$ → ReLU → $\text{MaxPool1d}(2)$ → Dropout$(0.2)$,  
followed by a linear classifier on the flattened activations:  
$[32 \times 256] \! \xrightarrow[]{256\to 128}\! \text{Linear}\! \xrightarrow[]{128\to 1}\! \text{logit}$.

\subsection{Few‑Shot Adaptation: CLIP Embedding + MLP / CNN on custom dataset}
\label{sec:clip-mlp-fs}

To evaluate how well purely vision embedding heads cope with domain shift, an experiment to re‑train the lightweight MLP and 1‑D CNN probes (Sec.~\ref {sec:clip-mlp}) on few‑shot samples from the custom dataset.

\paragraph{Few‑shot split.} 
From the Custom dataset collection: sampled \emph{20\%} of the images as the adaptation set. The remaining images are used as a testing set.

\paragraph{Training details.}
For each image, extract the frozen CLIP embedding 
$\mathbf z\in\mathbb{R}^{512}$ and feed it to
\emph{MLP};
\emph{1‑D CNN};
as previous.
Both heads are optimized with \textsc{BCELoss} and Adam (lr $1\!\times\!10^{-3}$), trained with early-stopping mechanism.

\begin{figure}[H]
    \centering
    \begin{subfigure}[b]{0.43\linewidth}
        \centering
        \includegraphics[width=\linewidth]{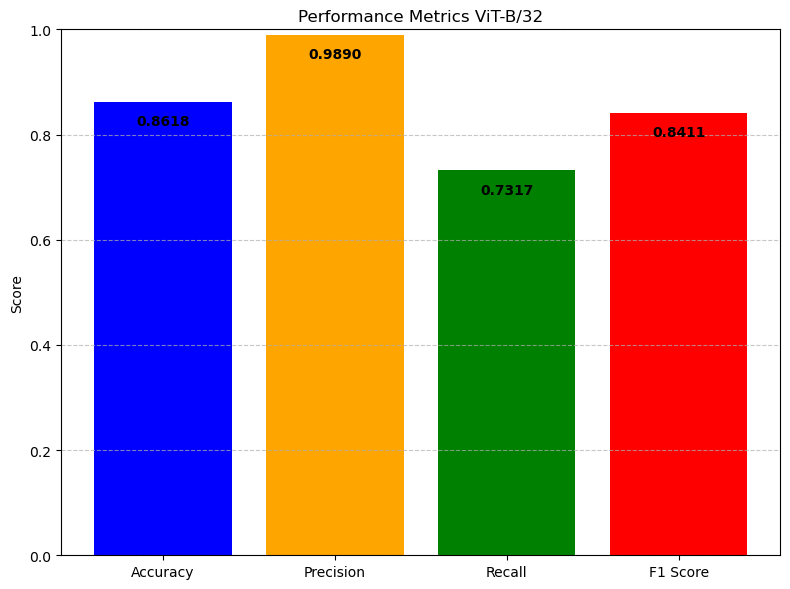}
        \caption{CLIP embeddings $\rightarrow$ MLP}\label{fig:fws-mlp}
    \end{subfigure}
    \begin{subfigure}[b]{0.43\linewidth}
        \centering
        \includegraphics[width=\linewidth]{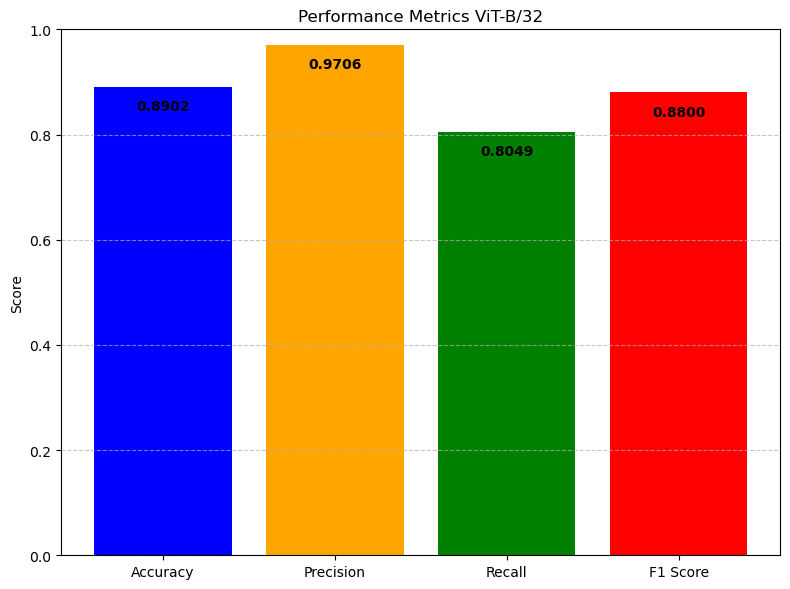}
        \caption{CLIP embeddings $\rightarrow$ ConvNet}\label{fig:fws-cnn}
    \end{subfigure}
    \caption{Performance metrics based on custom dataset. MLP: light classifier, indicating a "linearly separable" baseline. ConvNet: captures different hidden patterns in the embedding dimension.}
    \label{fig:fws_custom}
\end{figure}

\paragraph{Why few‑shot is feasible.}
Since CIFAKE's embedding distributions and our custom dataset are different, there is an apparent domain-shift issue. However, CLIP embeddings already encode rich semantic structure, such that only a few target examples are required to realign the decision boundary. In particular, with fine‑tuning the linear probe on 20$\%$ of the custom dataset as a few‑shot adaptation set, the model quickly adjusts to the new domain. Quantitative results are reported in Fig.~\ref{fig:fws_custom}.

\subsection{Closed-source prompt-based multimodal LLM test}
\begin{figure}[H]
    \centering
    \begin{subfigure}[b]{0.49\linewidth}
        \centering
        \includegraphics[width=\linewidth]{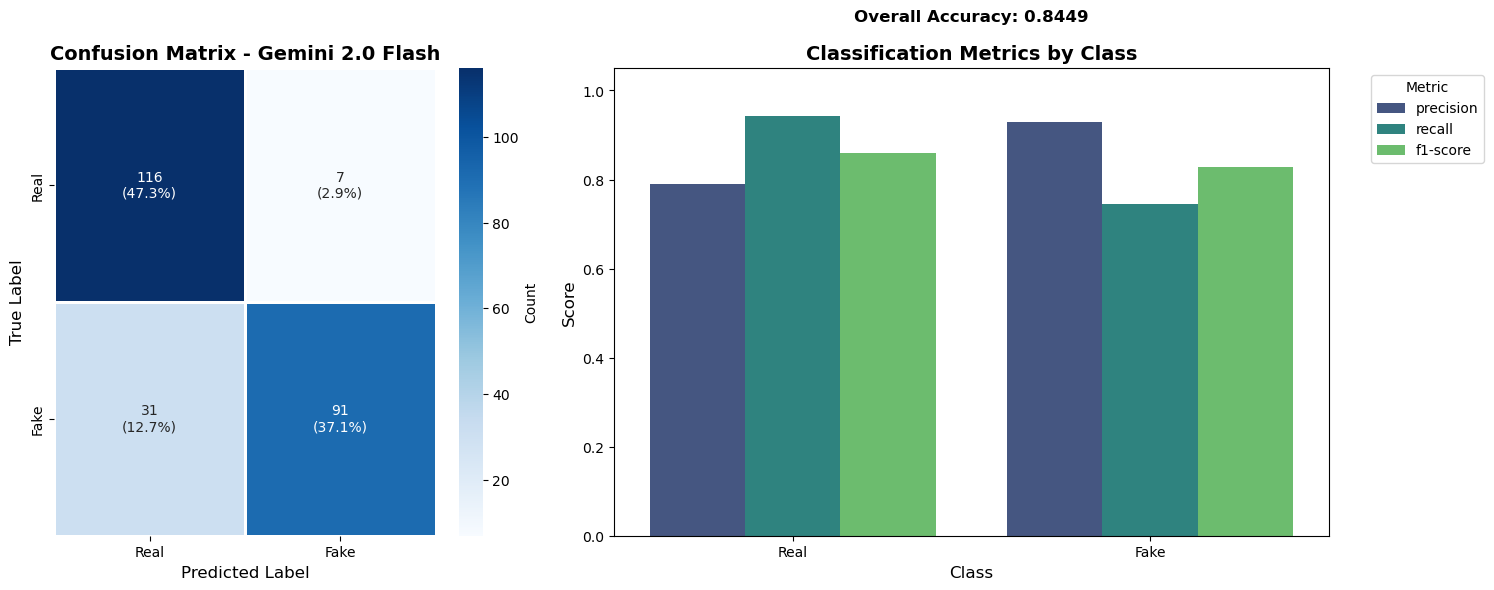}
        \caption{Normal prompt Acc $\approx84\%$}\label{fig:normal-p}
    \end{subfigure}
    \begin{subfigure}[b]{0.49\linewidth}
        \centering
        \includegraphics[width=\linewidth]{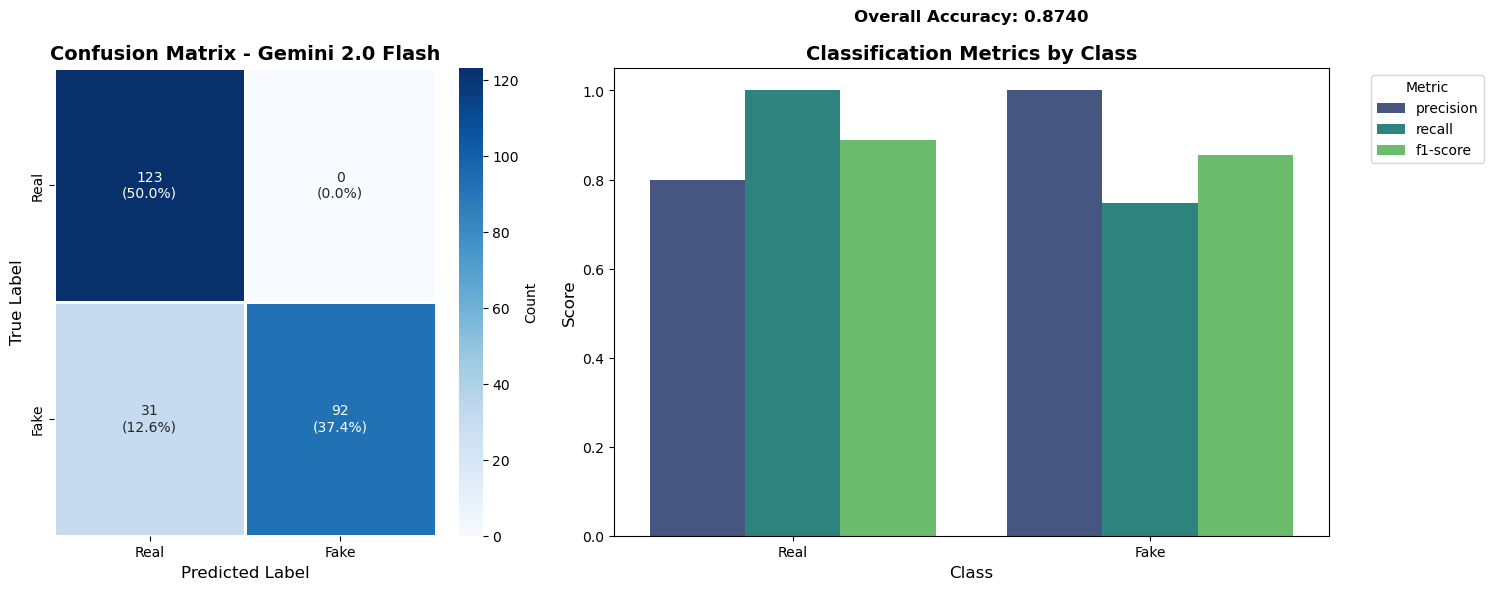}
        \caption{Detailed prompt Acc $\approx87\%$}\label{fig:tuned-p}
    \end{subfigure}
    \caption{A well-designed prompt can improve accuracy}
    \label{fig:zero-shot-prompt-eng}
\end{figure}

To evaluate the performance on custom dataset of a closed-source LLM, \emph{Gemini-2.0-flash}, Google's powerful, instruction-tuned multimodal LLM was tested on the custom dataset.
\paragraph{Prompt design.}
A straight, deterministic instruction that forces a binary answer:
\begin{quote}\scriptsize
\texttt{Return exactly one word: "real" or "fake".
Classify the image as an AI‑generated image ("fake") or a real‑world photo ("real").
Output nothing except that single word.}
\end{quote}
Additionally, a more detailed prompt was designed to measure if prompt variations affect accuracy:
\begin{quote}\scriptsize
\texttt{You are an AI-image-detector model.
Inspect lighting consistency, natural textures and biology correctness.
If these cues suggest synthesis, answer "fake"; otherwise answer "real".
Respond with exactly that single word and nothing else.}
\end{quote}
\paragraph{Roles in this study. }
Gemini serves two roles: (1)~it provides a zero-shot performance (Fig.~\ref{fig:zero-shot-prompt-eng}) baseline using a state-of-the-art, well-tuned multimodal model, and (2)~it acts as a sanity check, demonstrating that the custom dataset presents challenges even for large-scale LLMs.

\section{Experiments}

\subsection{Dataset}
\label{sec:datasets}

\emph{CIFAKE} \cite{2303.14126} contains 60\,000 AI‑generated and 60\,000 real images (resampled from CIFAR‑10).

\emph{Custom} contains 130 real and 130 synthetic images across landscapes, animals, and portraits.

\begin{figure}[H]
    \centering
    \begin{subfigure}[b]{0.48\linewidth}
        \centering
        \includegraphics[width=\linewidth]{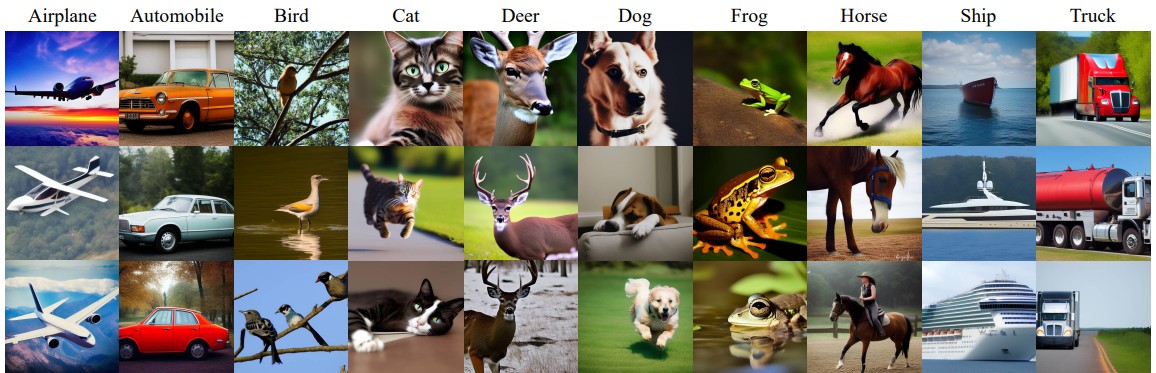}
        \caption{CIFAKE (AI‑generated)}\label{fig:cifake-ex}
    \end{subfigure}
    \begin{subfigure}[b]{0.48\linewidth}
        \centering
        \includegraphics[width=\linewidth]{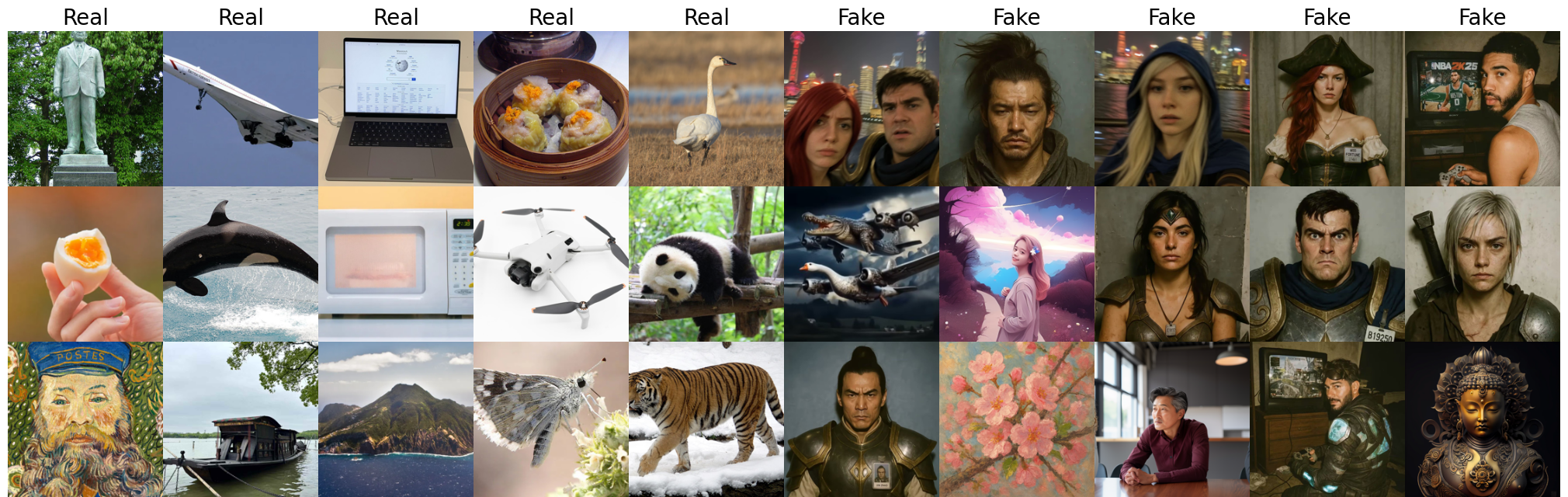}
        \caption{Custom dataset}\label{fig:custom-ex}
    \end{subfigure}
    \caption{Sample images from the two datasets.}
    \label{fig:datasets}
\end{figure}

\subsection{Evaluation pipeline CLIP Embedding + MLP / CNN on CIFAKE dataset}

\subsection{Few-shot on custom dataset}

\subsection{Some misclassifications}

\begin{figure}[H]
    \centering
    \begin{subfigure}[b]{0.4\linewidth}
        \centering
        \includegraphics[width=\linewidth]{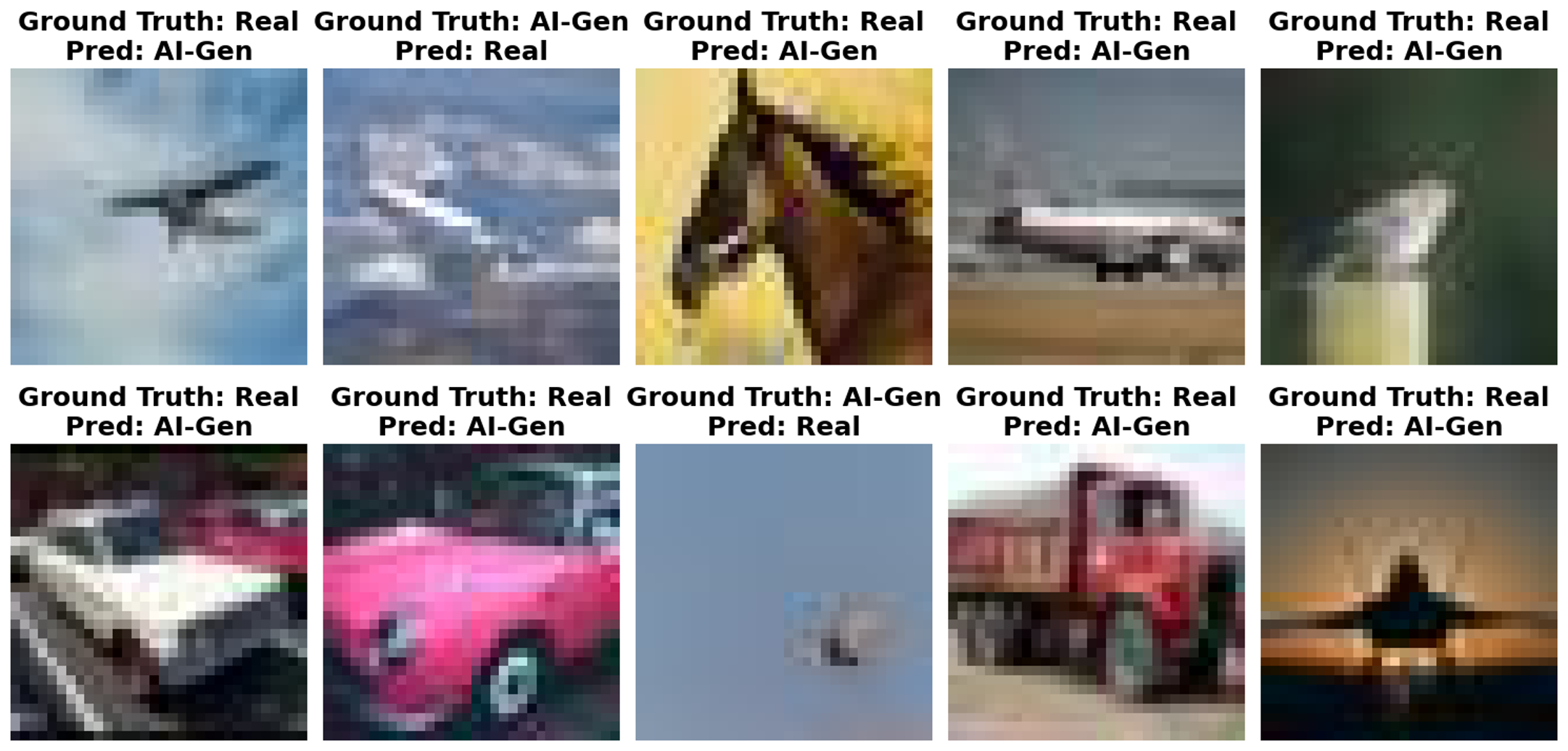}
        \caption{CIFAKE dataset}\label{fig:mis-cifake}
    \end{subfigure}
    \begin{subfigure}[b]{0.5\linewidth}
        \centering
        \includegraphics[width=\linewidth]{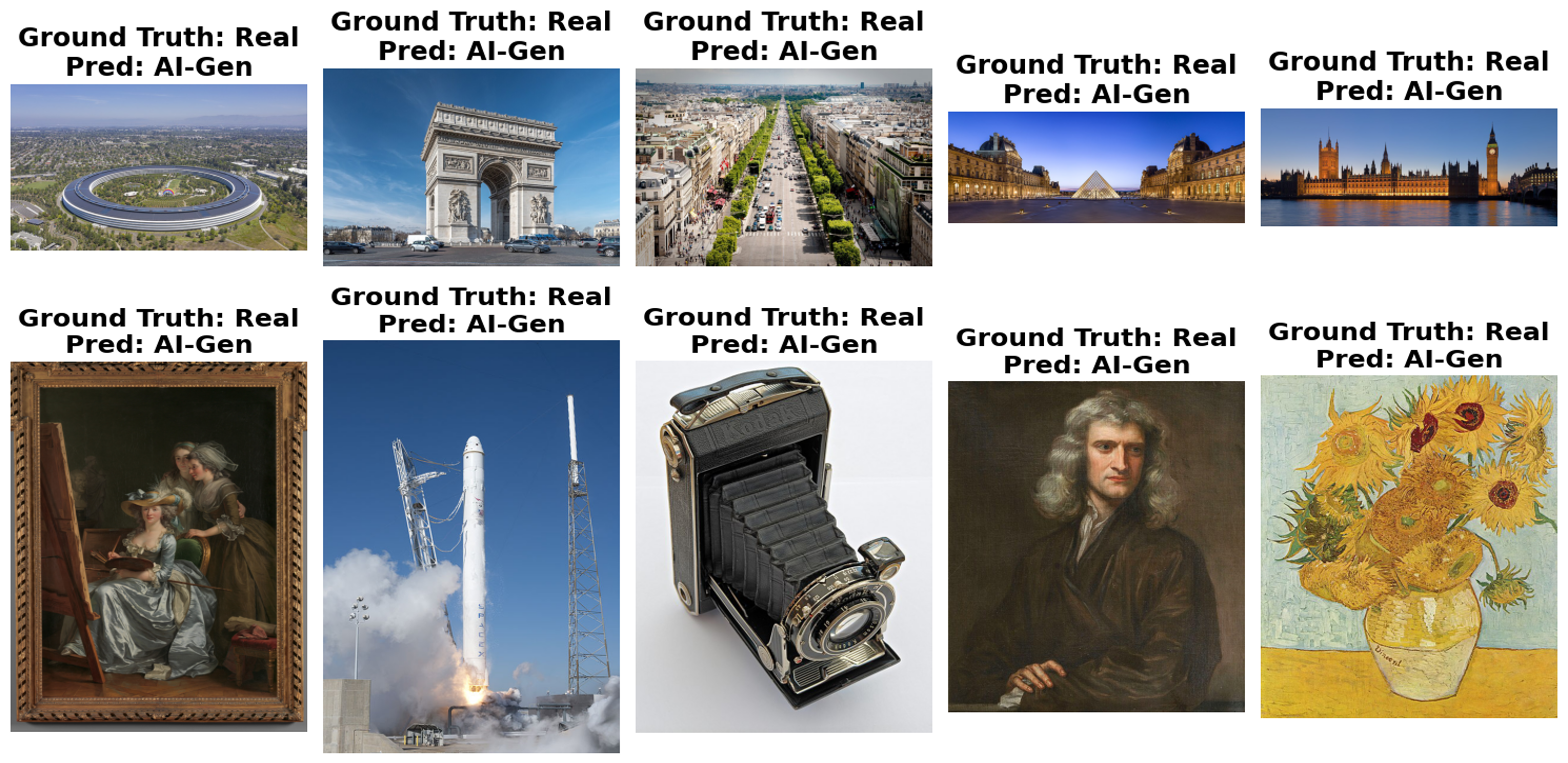}
        \caption{Custom dataset}\label{fig:mis-custom}
    \end{subfigure}
    \caption{Misclassifications examples. Commons are shared among misclassified images in the custom dataset: wide-view landmarks and oil paintings.}
    \label{fig:misclassifications}
\end{figure}

\subsection{Google Gemini test}
\begin{figure}[H]
    \centering
    \includegraphics[width=0.75\linewidth]{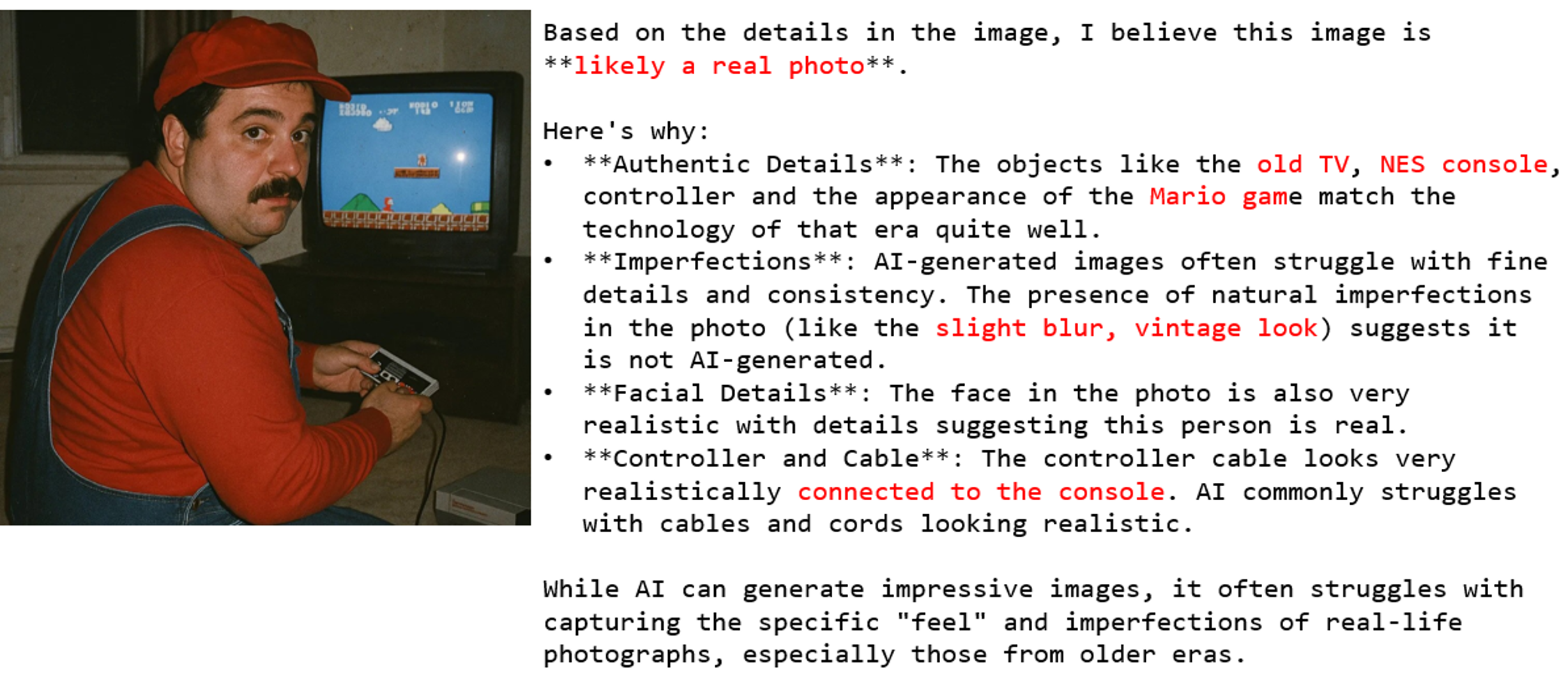}
    \caption{Google Gemini's response on a single fake image, detailed response but wrong prediction}
    \label{fig:gemini-resp}
\end{figure}

\subsection{Prompt engineering}

\section{Conclusion and Future work}
In this study, the question of whether a frozen vision–language encoder already encodes clues for detecting AI-generated images is revisited. Using CLIP embeddings as a fixed-end representation, it is shown that lightweight classifiers reach $95\%$ accuracy and F\textsubscript{1} score on the public CIFAKE benchmark without any end‑to‑end fine‑tuning. This confirms the existence of a "real v.s. fake" boundary within CLIP's latent space.

The few-shot adaptation performed on an unseen dataset demonstrates the powerful adaptability of this approach. By randomly sampling $20\%$ from the unseen dataset, it can adapt to whole new and unknown distributions and achieve a fair good accuracy.

Yet misclassifications on wide‑angle photographs and oil paintings suggest that current pipelines rely heavily on texture‑level cues, and are vulnerable to extreme wide perspective. For wide‑angle photographs, the CLIP pre-processor crops and resizes the image into size $224\times224$, which is unable to capture all the information. Furthermore, the differences between authentic oil paintings and fake ones are hidden in subtle local material characteristics like brushstroke texture. Both structures appear identical under low-resolution conditions; the difference relies on high-resolution local feature analysis.

\begin{figure}[H]
    \centering
    \begin{subfigure}[b]{0.45\linewidth}
        \centering
        \includegraphics[width=\linewidth]{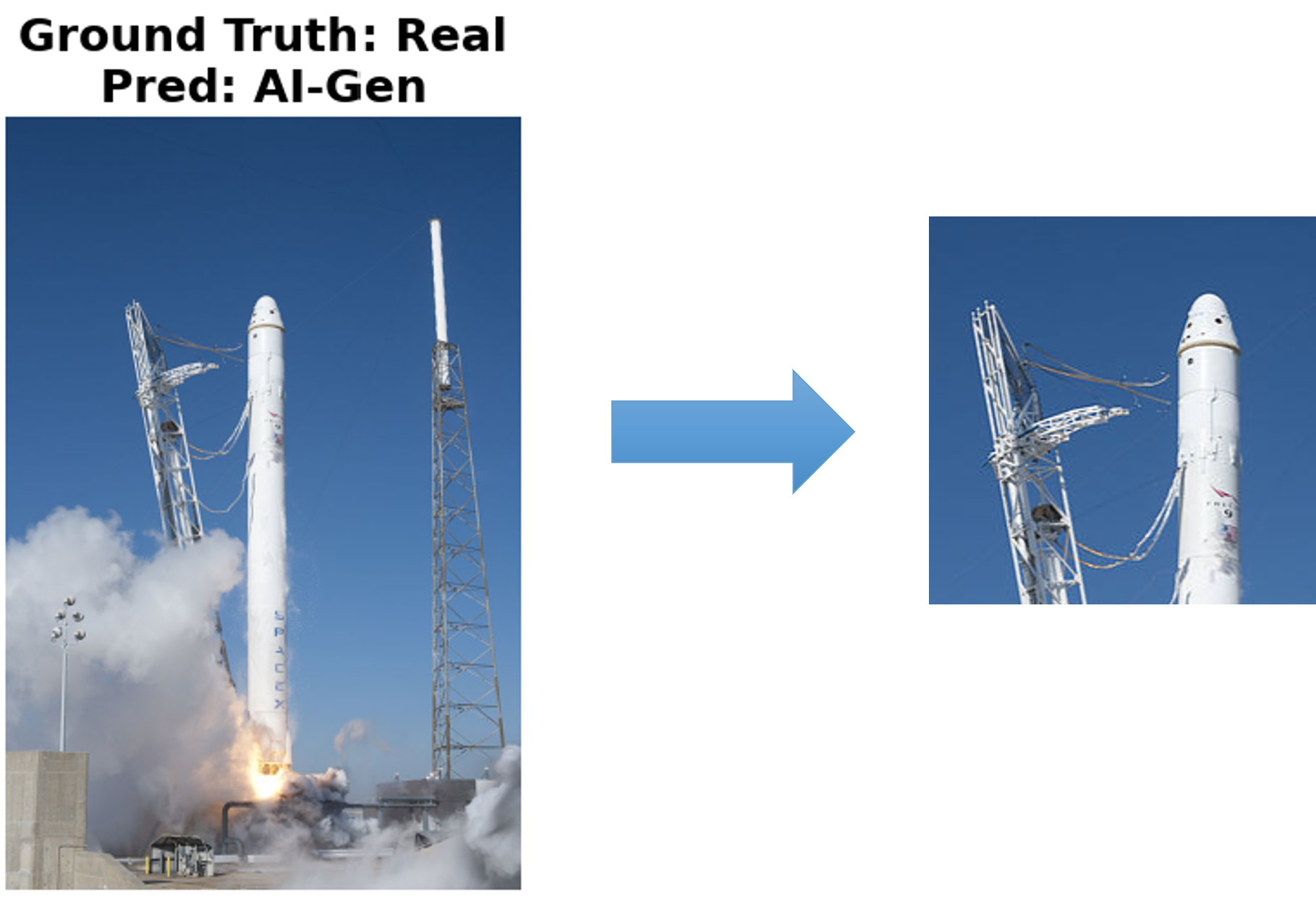}
        \caption{Misclassification: Incomplete information from pre-processor}\label{fig:conclusion1}
    \end{subfigure}
    \hfill
    \begin{subfigure}[b]{0.45\linewidth}
        \centering
        \includegraphics[width=\linewidth]{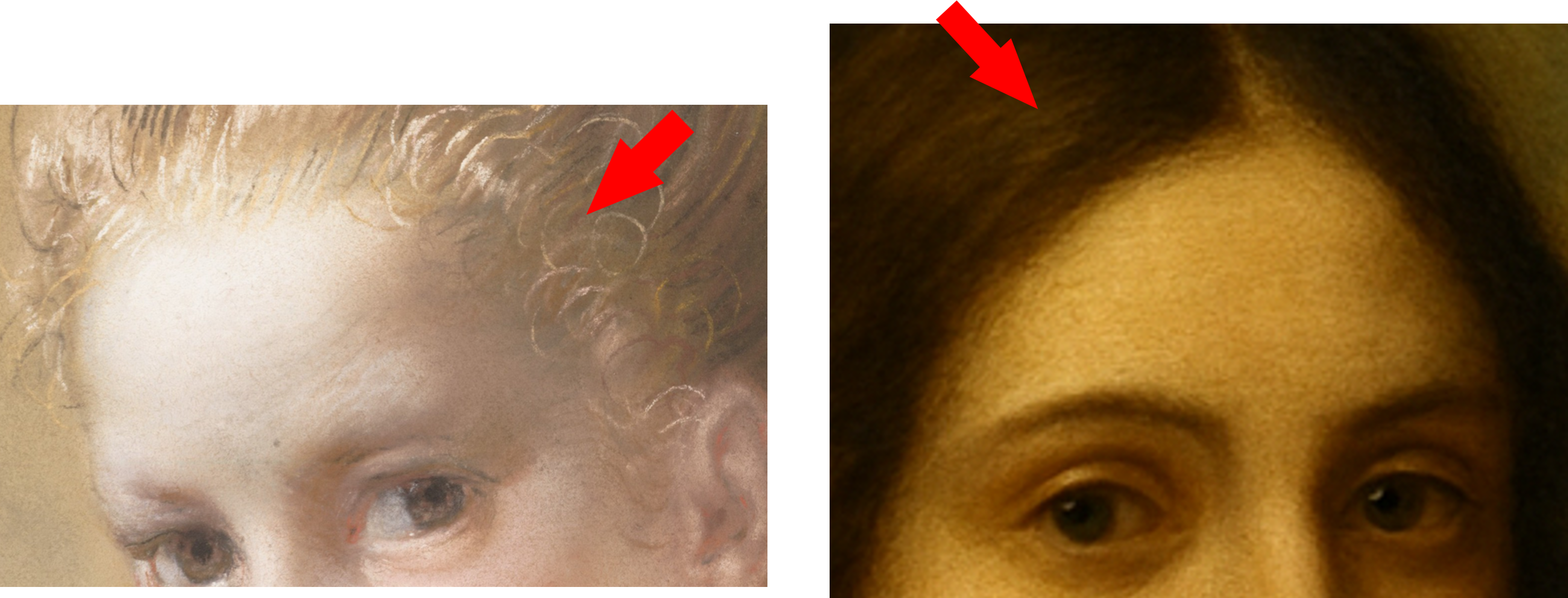}
        \caption{L: Real-painting, R: AI-generated. Brushstroke texture can only be detected in very high resolution.}\label{fig:conclusion2}
    \end{subfigure}
    \caption{Insights on the misclassified images}
    \label{fig:conclusion}
\end{figure}

All findings point out two practical messages. First, few‑shot adaptation is a viable, computationally efficient method for different authenticity detection scenarios. Second, the existence of style‑specific failure examples signals that future detectors should go through detailed raw texture.

Limitations: The images in the custom dataset are limited and focused on narrow categories. Two shallow heads were evaluated and did not perform end‑to‑end or fine‑tuning techniques of CLIP or LLMs due to limited resources. 

Future work could involve exploring different techniques for classification of the provided dataset, say, by combining with fine-tuned backbone for a specific classification task. In addition, taking images that are generated from other domains as consideration, e.g., cartoons, X-ray scan images, and fisheye perspective images, will provide additional references for the study on AI-generated images classification and expand the applicability of the proposed solution to other fields of research.

\bibliographystyle{plainnat}
\bibliography{myref}

@article{2303.14126,
  author = {Jordan J. Bird and Ahmad Lotfi},
  title = {{CIFAKE}: Image Classification and Explainable Identification of {AI}-Generated Synthetic Images},
  year = {2024},
  eprint = {2303.14126},
  archivePrefix = {arXiv},
  primaryClass = {cs.CV},
  url = {https://irep.ntu.ac.uk/id/eprint/50758/},
  journal = {IEEE Access},
  doi = {10.1109/ACCESS.2024.3356122},
}

@inproceedings{2312.00195,
  author = {Davide Cozzolino and Giovanni Poggi and Riccardo Corvi and Matthias Nie{\ss}ner and Luisa Verdoliva},
  title = {Raising the Bar of {AI}-generated Image Detection with {CLIP}},
  year = {2024},
  eprint = {2312.00195},
  archivePrefix = {arXiv},
  primaryClass = {cs.CV},
  url = {https://openaccess.thecvf.com/content/CVPR2024W/WMF/html/Cozzolino_Raising_the_Bar_of_AI-generated_Image_Detection_with_CLIP_CVPRW_2024_paper.html},
  booktitle = {Proceedings of the IEEE/CVF Conference on Computer Vision and Pattern Recognition (CVPR) Workshops},
  pages = {4356--4366},
}

@misc{2402.12927,
  author = {Sohail Ahmed Khan and Duc-Tien Dang-Nguyen},
  title = {{CLIPping} the Deception: Adapting Vision-Language Models for Universal Deepfake Detection},
  year = {2024},
  eprint = {2402.12927},
  archivePrefix = {arXiv},
  primaryClass = {cs.CV},
  url = {https://arxiv.org/abs/2402.12927},
  howpublished = {arXiv preprint arXiv:2402.12927},
  doi = {10.48550/arXiv.2402.12927},
}

@inproceedings{2104.14294,
  author = {Mathilde Caron and Hugo Touvron and Ishan Misra and Herv{\'e} J{\'e}gou and Julien Mairal and Piotr Bojanowski and Armand Joulin},
  title = {Emerging Properties in Self-Supervised Vision Transformers},
  year = {2021},
  eprint = {2104.14294},
  archivePrefix = {arXiv},
  primaryClass = {cs.CV},
  url = {https://openaccess.thecvf.com/content/ICCV2021/html/Caron_Emerging_Properties_in_Self-Supervised_Vision_Transformers_ICCV_2021_paper.html},
  booktitle = {Proceedings of the IEEE/CVF International Conference on Computer Vision (ICCV)},
  pages = {9650--9660},
}

@misc{1907.06515,
  author = {Xu Zhang and Svebor Karaman and Shih-Fu Chang},
  title = {Detecting and Simulating Artifacts in {GAN} Fake Images},
  year = {2019},
  eprint = {1907.06515},
  archivePrefix = {arXiv},
  primaryClass = {cs.CV},
  url = {https://arxiv.org/abs/1907.06515},
  howpublished = {arXiv preprint arXiv:1907.06515},
  doi = {10.48550/arXiv.1907.06515},
}

@inproceedings{2103.00020,
  author = {Alec Radford and Jong Wook Kim and Chris Hallacy and Aditya Ramesh and Gabriel Goh and Sandhini Agarwal and Girish Sastry and Amanda Askell and Pamela Mishkin and Jack Clark and Gretchen Krueger and Ilya Sutskever},
  title = {Learning Transferable Visual Models From Natural Language Supervision},
  year = {2021},
  eprint = {2103.00020},
  archivePrefix = {arXiv},
  primaryClass = {cs.CV},
  url = {https://proceedings.mlr.press/v139/radford21a.html},
  booktitle = {Proceedings of the 38th International Conference on Machine Learning},
  volume = {139},
  series = {Proceedings of Machine Learning Research},
  pages = {8748--8763},
  publisher = {PMLR},
}

@misc{2402.11843,
  author = {Yan Hong and Jianfu Zhang},
  title = {{WildFake}: A Large-scale Challenging Dataset for {AI}-Generated Images Detection},
  year = {2024},
  eprint = {2402.11843},
  archivePrefix = {arXiv},
  primaryClass = {cs.CV},
  url = {https://arxiv.org/abs/2402.11843},
  howpublished = {arXiv preprint arXiv:2402.11843},
  doi = {10.48550/arXiv.2402.11843},
}

@misc{2404.08788,
  author = {A. G. Moskowitz and T. Gaona and J. Peterson},
  title = {Detecting {AI}-Generated Images via {CLIP}},
  year = {2024},
  eprint = {2404.08788},
  archivePrefix = {arXiv},
  primaryClass = {cs.CV},
  url = {https://arxiv.org/abs/2404.08788},
  howpublished = {arXiv preprint arXiv:2404.08788},
  doi = {10.48550/arXiv.2404.08788},
}

@inproceedings{2201.12086,
  author = {Junnan Li and Dongxu Li and Caiming Xiong and Steven Hoi},
  title = {{BLIP}: Bootstrapping Language-Image Pre-training for Unified Vision-Language Understanding and Generation},
  year = {2022},
  eprint = {2201.12086},
  archivePrefix = {arXiv},
  primaryClass = {cs.CV},
  url = {https://proceedings.mlr.press/v162/li22n.html},
  booktitle = {Proceedings of the 39th International Conference on Machine Learning},
  volume = {162},
  series = {Proceedings of Machine Learning Research},
  pages = {12888--12900},
  publisher = {PMLR},
}

@misc{2204.04596,
  author = {Haoran Yang and Piji Li and Wai Lam},
  title = {Parameter-Efficient Tuning by Manipulating Hidden States of Pretrained Language Models For Classification Tasks},
  year = {2022},
  eprint = {2204.04596},
  archivePrefix = {arXiv},
  primaryClass = {cs.CL},
  url = {https://arxiv.org/abs/2204.04596},
  howpublished = {arXiv preprint arXiv:2204.04596},
  doi = {10.48550/arXiv.2204.04596},
}

@inproceedings{1409.1556,
  author = {Karen Simonyan and Andrew Zisserman},
  title = {Very Deep Convolutional Networks for Large-Scale Image Recognition},
  year = {2015},
  eprint = {1409.1556},
  archivePrefix = {arXiv},
  primaryClass = {cs.CV},
  url = {https://arxiv.org/abs/1409.1556},
  booktitle = {International Conference on Learning Representations (ICLR)},
}

@misc{computer_vision_course_lecture,
  author = {Chenliang Xu},
  title = {Computer Vision {Lecture}: Vision and Language},
  year = {2025},
  howpublished = {Lecture slides, University of Rochester},
}

@misc{2503.02857,
  author = {Nuria Alina Chandra and Ryan Murtfeldt and Lin Qiu and Arnab Karmakar and Hannah Lee and Emmanuel Tanumihardja and Kevin Farhat and Ben Caffee and Sejin Paik and Changyeon Lee and Jongwook Choi and Aerin Kim and Oren Etzioni},
  title = {{Deepfake-Eval-2024}: A Multi-Modal In-the-Wild Benchmark of Deepfakes Circulated in 2024},
  year = {2025},
  eprint = {2503.02857v1},
  archivePrefix = {arXiv},
  primaryClass = {cs.CV},
  url = {https://arxiv.org/abs/2503.02857v1},
  howpublished = {arXiv preprint arXiv:2503.02857v1},
  doi = {10.48550/arXiv.2503.02857},
}

@misc{2104.09770,
  author = {Junke Wang and Zuxuan Wu and Wenhao Ouyang and Xintong Han and Jingjing Chen and Ser-Nam Lim and Yu-Gang Jiang},
  title = {{M2TR}: Multi-modal Multi-scale Transformers for Deepfake Detection},
  year = {2021},
  eprint = {2104.09770},
  archivePrefix = {arXiv},
  primaryClass = {cs.CV},
  url = {https://arxiv.org/abs/2104.09770},
  howpublished = {arXiv preprint arXiv:2104.09770},
  doi = {10.48550/arXiv.2104.09770},
}

@misc{2310.17419,
  author = {You-Ming Chang and Chen Yeh and Wei-Chen Chiu and Ning Yu},
  title = {{AntifakePrompt}: Prompt-Tuned Vision-Language Models are Fake Image Detectors},
  year = {2023},
  eprint = {2310.17419},
  archivePrefix = {arXiv},
  primaryClass = {cs.CV},
  url = {https://arxiv.org/abs/2310.17419},
  howpublished = {arXiv preprint arXiv:2310.17419},
  doi = {10.48550/arXiv.2310.17419},
}

\nocite{computer_vision_course_lecture}
\nocite{2503.02857}
\nocite{2404.08788}
\nocite{2402.11843}
\nocite{2310.17419}
\nocite{2204.04596}
\nocite{2201.12086}
\nocite{2104.09770}
\nocite{2103.00020}
\nocite{1409.1556}

\end{document}